\documentclass[letterpaper]{article} 
\usepackage{aaai25}  
\usepackage{times}  
\usepackage{helvet}  
\usepackage{courier}  
\usepackage[hyphens]{url}  
\usepackage{graphicx} 
\usepackage{natbib}  
\usepackage{caption} 
\usepackage{algorithm}
\usepackage{algorithmic}
\usepackage{amssymb}
\usepackage{booktabs}
\usepackage{multirow}
\usepackage{soul}
\usepackage{array}

\usepackage{newfloat}
\usepackage{listings}
\DeclareCaptionStyle{ruled}{labelfont=normalfont,labelsep=colon,strut=off} 
\floatstyle{ruled}
\newfloat{listing}{tb}{lst}{}
\floatname{listing}{Listing}
\title{Just a Few Glances: Open-Set Visual Perception with Image Prompt Paradigm}
\author{
    Jinrong Zhang\textsuperscript{\rm 2}\thanks{The work was done while Jinrong Zhang was a research intern at Xiaomi AI Lab.},
    Penghui Wang\textsuperscript{\rm 1},
    Chunxiao Liu\textsuperscript{\rm 1},
    Wei Liu\textsuperscript{\rm 1},
    Dian Jin\textsuperscript{\rm 1},
    Qiong Zhang\textsuperscript{\rm 1}\equalcontrib\thanks{Corresponding author.},
    Erli Meng\textsuperscript{\rm 1}\footnotemark[3],
    Zhengnan Hu\textsuperscript{\rm 1}
}
\affiliations{
    \textsuperscript{\rm 1}Xiaomi AI Lab, Beijing, China\\
    \textsuperscript{\rm 2}Dalian University of Technology, Dalian, China\\

    zjr15272565639@mail.dlut.edu.cn, \{wangpenghui, liuchunxiao, liuwei67, jindian1, zhangqiong1, mengerli, huzhengnan\}@xiaomi.com
}

\usepackage{bibentry}

\begin{document}

\maketitle

\begin{abstract}
    To break through the limitations of pre-training models on fixed categories, Open-Set Object Detection (OSOD) and Open-Set Segmentation (OSS) have attracted a surge of interest from researchers. Inspired by large language models, mainstream OSOD and OSS methods generally utilize text as a prompt, achieving remarkable performance. Following SAM paradigm, some researchers use visual prompts, such as points, boxes, and masks that cover detection or segmentation targets. Despite these two prompt paradigms exhibit excellent performance, they also reveal inherent limitations. On the one hand, it is difficult to accurately describe characteristics of specialized category using textual description. On the other hand, existing visual prompt paradigms heavily rely on multi-round human interaction, which hinders them being applied to fully automated pipeline. To address the above issues, we propose a novel prompt paradigm in OSOD and OSS, that is, \textbf{Image Prompt Paradigm}. This brand new prompt paradigm enables to detect or segment specialized categories without multi-round human intervention. To achieve this goal, the proposed image prompt paradigm uses just a few image instances as prompts, and we propose a novel framework named \textbf{MI Grounding} for this new paradigm. In this framework, high-quality image prompts are automatically encoded, selected and fused, achieving the single-stage and non-interactive inference. We conduct extensive experiments on public datasets, showing that MI Grounding achieves competitive performance on OSOD and OSS benchmarks compared to text prompt paradigm methods and visual prompt paradigm methods. Moreover, MI Grounding can greatly outperform existing method on our constructed specialized ADR50K dataset.
\end{abstract}

%

\section{Introduction}

To break through the limitations of pre-training models on fixed categories, Open-Set Object Detection (OSOD) and Open-Set Segmentation (OSS) have attract a surge of interest from researchers. In these fields, trained models can not only detect or segment predefined specific categories but also generalize to open scenarios, which greatly improve the ability and applicability~\cite{li2022grounded}.

Inspired by the remarkable success achieved by foundational models~\cite{radford2021learning,li2022grounded}, mainstream OSOD and OSS methods employ a prompt as an input, which tells the model what to detect or segment in the image. Existing prompt paradigms can be mainly categorized into two types: text prompt paradigm and visual prompt paradigm. As for text prompt paradigm, users are required to provide a textual description to depict characteristics of detection or segmentation targets, and models are trained to align text prompt with visual contents in the latent space~\cite{liu2023grounding,ding2022open}. Following SAM~\cite{kirillov2023segment}, another line of approaches employ visual prompts, such as points, boxes, and masks. The visual prompt needs to be manually designed that can locate specific targets. Such a design makes this process generally involves multi-round interaction to avoid ambiguous prompts~\cite{kirillov2023segment}.

\begin{figure}[t]
    \centering
    \includegraphics[width=1.\linewidth]{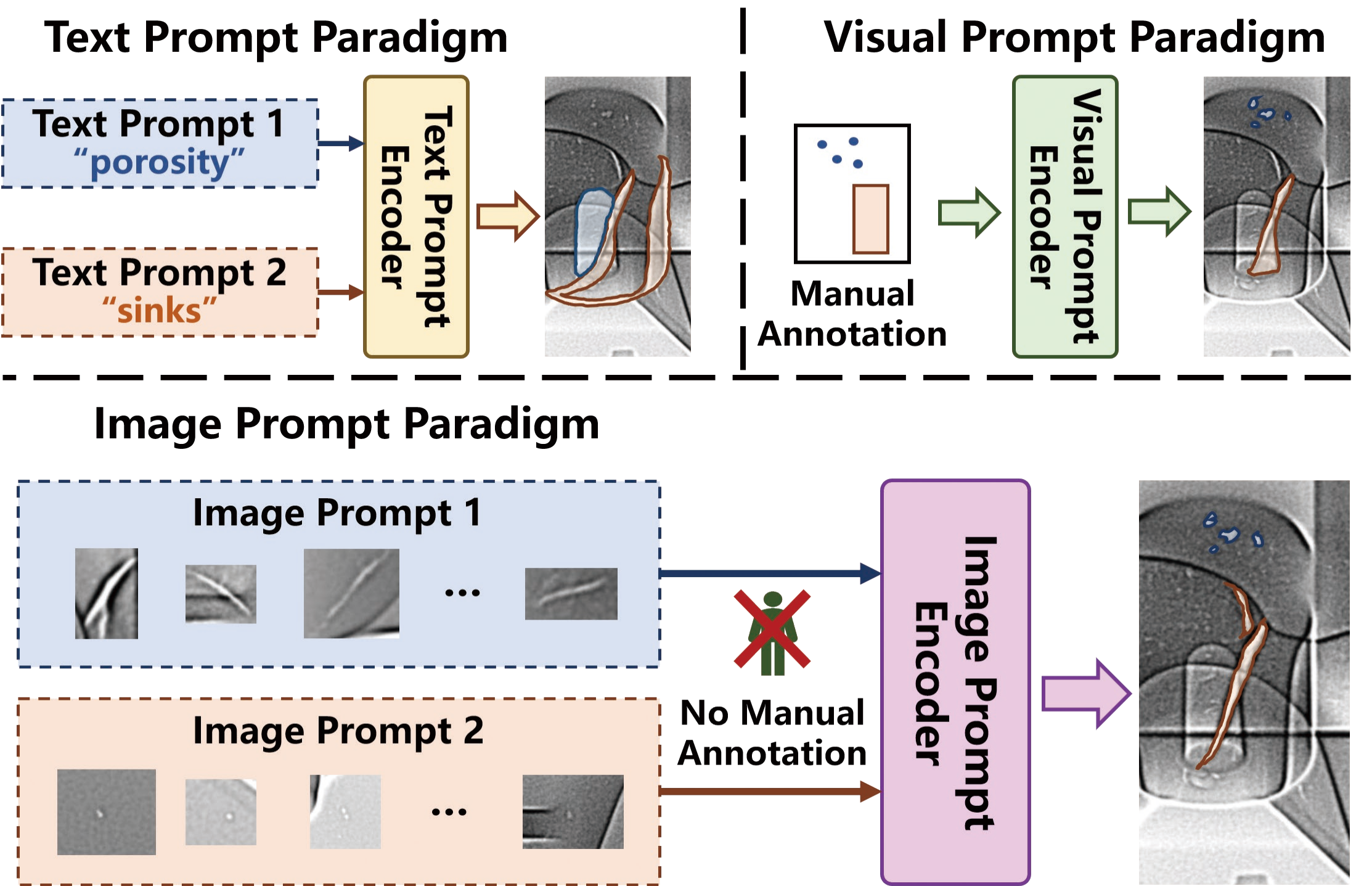}
    \caption{Image prompt paradigm vs. previous prompt paradigms. The text prompt paradigm struggles to accurately describe specialized categories. The visual prompt paradigm relies on multi-round human interaction. The proposed image prompt paradigm uses just a few image instances which can handle specialized categories without any manual annotation.}
    \label{fg_image_prompt_paradigm}
\end{figure}

However, textual and visual prompt paradigms have the following limitations. First of all, the visual feature of specialized categories are difficult to be accurately described by text, and hence hinder the application of text prompt paradigm~\cite{li2024visual,jiang2024t}. Second, visual prompts heavily rely on multi-round human interaction, which makes it difficult to be applied into production pipelines~\cite{kirillov2023segment}. As shown in Figure.1, in X-ray defect detection, we need to detect and segment specialized categories, such as ``shrinkage porosity", ``sinks", and ``porosity". These concepts are specific to the X-ray field, which cannot reflect the visual characteristics without industrial knowledge, such as their shape, size, and texture etc. Visual prompt might alleviate this issue by providing bounding boxes of ``shrinkage porosity", ``sinks", and ``porosity" as prompts, but it requires users to annotate or check bounding boxes to make sure they cover the target areas~\cite{kirillov2023segment}. These interaction processes make a single-stage, fully automated inference pipeline impossible.

In this work, we establish a novel visual perception paradigm, i.e. \textbf{Image Prompt Paradigm}, which completely abandons traditional text prompts and visual prompts, achieving a single-stage and fully automated inference. Inspired by the fact that humans can quickly grasp the characteristics of a specific category after taking \textbf{just a few glances} at its instances, the proposed image prompt paradigm utilizes just a few image instances of target as prompts. These instances are automatically constructed and calculated by our proposed \textbf{MI Grounding} framework, which uses \textbf{m}ultiple \textbf{i}mages as prompts. To bridge the gap between specialized categories and visual content, MI Grounding introduce an image prompts selection encoder, which can encode, select and integrate image prompts. The encoder module possesses extensive prior knowledge at the visual level, and can extract inherent distinctive semantic information of image prompts. The encoded image prompts are then selected and integrated to highlight high-quality image prompts automatically. After aligning the image prompts with the predicted objects, MI Grounding is learned to handle specialized categories that are difficult to describe using text, and achieves single-stage and non-interactive inference. Extensive experiments have shown that the proposed image prompt paradigm and MI Grounding achieve excellent detection and segmentation performance.

Concretely, our contributions can be summarized as follows:

\begin{itemize}
    \item We propose a novel visual perception paradigm: Image Prompt Paradigm. Different from existing text and visual prompt paradigm, this paradigm uses just a few image instances as prompts, which can understand specialized categories that are hard to describe by text in a single-stage and non-interactive manner.
    \item We propose a novel framework named MI Grounding tailored for the proposed image prompt paradigm. MI Grounding utilizes just a few image prompts to perform Open-Set Object Detection and Open-Set Segmentation, and propose an image prompt selection encoder to select and integrate high-quality prompts.
    \item Our approach achieves competitive performance on several datasets compared with mainstream Open-Set Object Detection and Open-Set Object Segmentation methods, which show the effectiveness of our proposed image prompt paradigm. To further demonstrate the superiority, we constructed a specialized ADR50K dataset, which contains a rich set of X-ray defect detection data. Experiments demonstrate that our approach can greatly improve the performance on this specialized dataset.
\end{itemize}

\begin{figure*}[t]
    \centering
    \includegraphics[width=1.\linewidth]{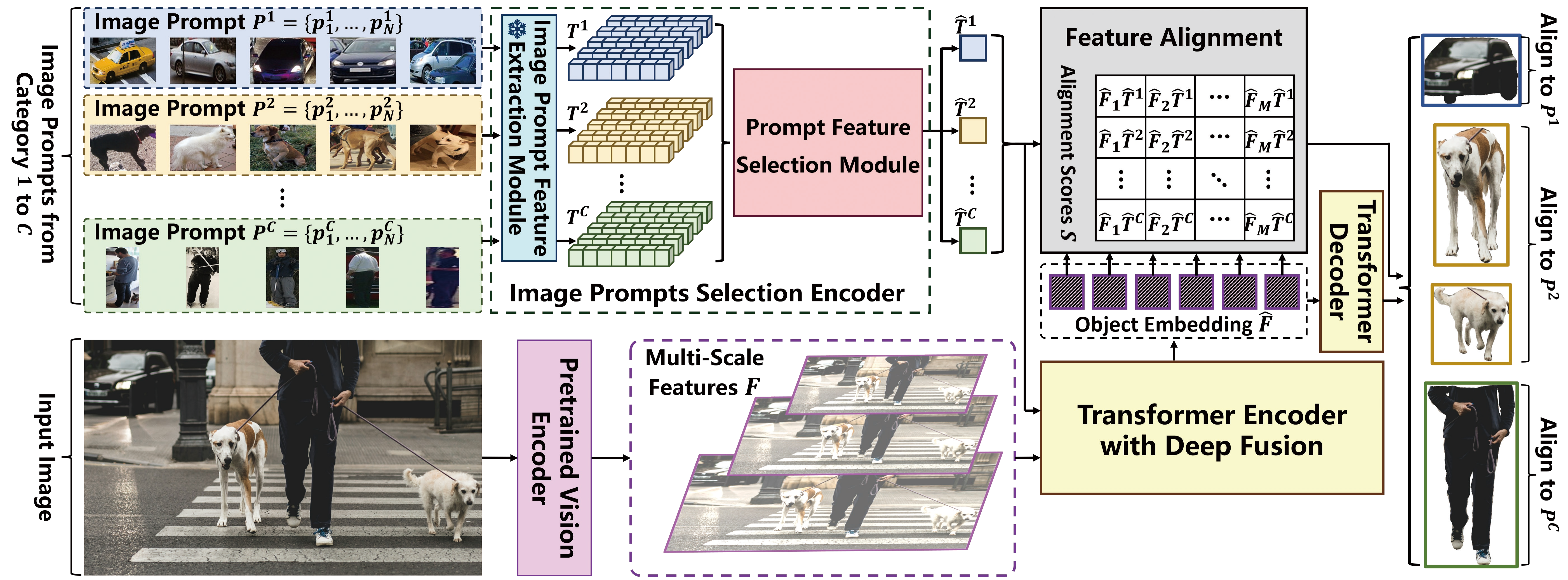}
    \caption{The overall framework of MI Grounding. Image prompts are encoded, selected, and integrated through the image prompts selection encoder (IPS encoder) to obtain category-specific prompt features. These prompt features are then deeply fused and aligned with multi-scale features from the input image to achieve open-set visual perception.}
    \label{fg_migrounding}
\end{figure*}

\section{Related Works}
\textbf{Visual Perception Based on Text Prompt Paradigm.} With the widespread application of foundational methods like CLIP~\cite{radford2021learning} and BERT~\cite{devlin2018bert}, open-vocabulary object detection and segmentation methods have achieved remarkable success in the general visual perception field. Researchers find that object detection and segmentation can be expressed as an alignment between text prompts and visual context information. Based on the concept above, these methods have made significant breakthroughs in zero-shot and few-shot learning. Grounding DINO~\cite{liu2023grounding} extends the training strategy of GLIP~\cite{li2022grounded} to DINO~\cite{zhang2022dino}, achieving strong open-set detection capabilities. DetCLIP~\cite{yao2022detclip} and RegionCLIP~\cite{zhong2022regionclip} utilize image-text pairs with pseudo boxes to expand region knowledge, thereby improving open-set performance. These text prompt paradigm methods rely on text encoders, like BERT, to model text queries. However, due to the ambiguity caused by the high information density of text and the potential mismatch between text descriptions and complex visual scenes, visual perception methods based on the text prompt paradigm have inherent limitations~\cite{li2024visual}.

\noindent \textbf{Visual Perception Based on Visual Prompt Paradigm.} Researchers have begun exploring alternative prompt paradigms. SAM~\cite{kirillov2023segment} pioneer an interactive open-set segmentation approach, introducing a novel prompt paradigm that includes boxes, points, masks, or lines covering the target. Subsequent researchers define this approach as the visual prompt paradigm and further explore its potential~\cite{li2024visual}. Semantic-SAM~\cite{li2023semantic} achieves semantic awareness by training on decoupled objects and parts classifications integrated from multiple datasets. Painter~\cite{wang2023images} and SegGPT~\cite{wang2023seggpt} adopt a generalist strategy to tackle diverse segmentation tasks, conceptualizing segmentation as an in-context coloring problem. DINOv~\cite{li2024visual} proposes a general contextual visual prompt framework, using visual context to understand new categories.

\noindent \textbf{Visual Perception with Image and Text Prompt.} To further enhance model performance, some researchers have introduced additional target images to augment text prompts. MQ-Det~\cite{xu2024multi} uses cross-attention and weighted addition to integrate image features into the text prompt, significantly improving model performance. However, when MQ-Det uses only images as prompts, the model's performance is poor. This indicates that MQ-Det has not fully exploited the potential of image prompts. In such methods, image prompts merely enhance the prompt features rather than lead the inference process.

\section{Method}
Our goal is to establish an image prompt paradigm, where the model completes open-set detection and segmentation by just taking a few glances at images of objects similar to the detection target. We first introduce how image prompts are constructed during the training and testing process. Then, we detail our proposed MI Grounding, including the model design and training strategy.

\subsection{Image Prompt Paradigm}
\label{sec_image_prompt_paprdigm}

In order to eliminate the tedious interaction process similar to the visual prompt paradigm and ensure that the data of the detection target is not leaked, we build an image prompt library using the training split of the dataset. Specifically, we crop instance targets from the original images based on their detection box labels and store them categorized by their class labels. The detailed process of extracting image prompts from the original images is as follows:

\begin{equation}
    p = Crop(I,{L_{box}}) = I[y:y + {h_p},x:x + {w_p}],
    \label{eq_crop}
\end{equation}

\noindent where $I \in {\mathbb{R}^{3 \times H \times W}}$ represents the original large image, and $p \in {\mathbb{R}^{3 \times {h_p} \times {w_p}}}$ is the instance target image obtained by cropping. \(H\) and \(W\) are the height and width of the original image. ${L_{box}} = \{ x,y,{w_p},{h_p}\}$ is the bounding box label, with $w_p$ and $h_p$ being the width and height of the corresponding instance target's bounding box, and $x$ and $y$ being the coordinates of the top-left corner of the bounding box. To ensure that target data from the test set is not leaked and enhance the model's robustness, we use only instance target images cropped from the training set as image prompts during both training and testing.

\subsection{MI Grounding}

As shown in Fig. \ref{fg_migrounding}, MI Grounding consists of an image prompts selection encoder (IPS encoder), a vision encoder, a transformer encoder with deep fusion following GLIP~\cite{li2022grounded}, and a transformer decoder. The IPS encoder extracts, selects, and integrates features from the image prompts, while the vision encoder extracts features from the input image. In the image prompt paradigm, the model uses a set of instance images $P^c = \{p_1^c, \dots, p_N^c\}$ of a specified category $c$ as prompts. The goal of MI Grounding is to detect and segment objects of the corresponding category from the input image $I$ based on $P^c$.

\noindent \textbf{Image Prompts Selection Encoder.} The IPS encoder consists of an image prompt feature extraction module and a prompt feature selection module. As for image prompt, how to extract their features to handle specialized categories is a crucial problem. Inspired by the text prompt that using pre-trained text encoders to fully utilize the latent semantic information, we learn that the critical point is to extract features that can distinguish the detection target from other instances. As a result, we employ pre-trained ViT as image prompt feature extractor, since it show good clustering properties, indicating pre-trained ViT contains semantic information useful for open-set visual perception. In the image prompt feature extractor, we compute features for all prompt images $P^c$ and aggregate them into a prompt feature matrix:

\begin{equation}
    {T^c} = STACK(ViT(p_1^c, \ldots ,p_N^c)),
    \label{eq_stack}
\end{equation}

\noindent where $STACK( \cdot )$ stands for feature stacking along the prompt quantity dimension, and $ViT( \cdot )$ represents a frozen vision transformer backbone. $T^c \in \mathbb{R}^{N \times D}$ denotes the prompt feature matrix composed of $N$ image prompts for a specified category $c$, where $D$ represents the feature dimension.

\begin{figure}[t]
    \centering
    \includegraphics[width=1.\linewidth]{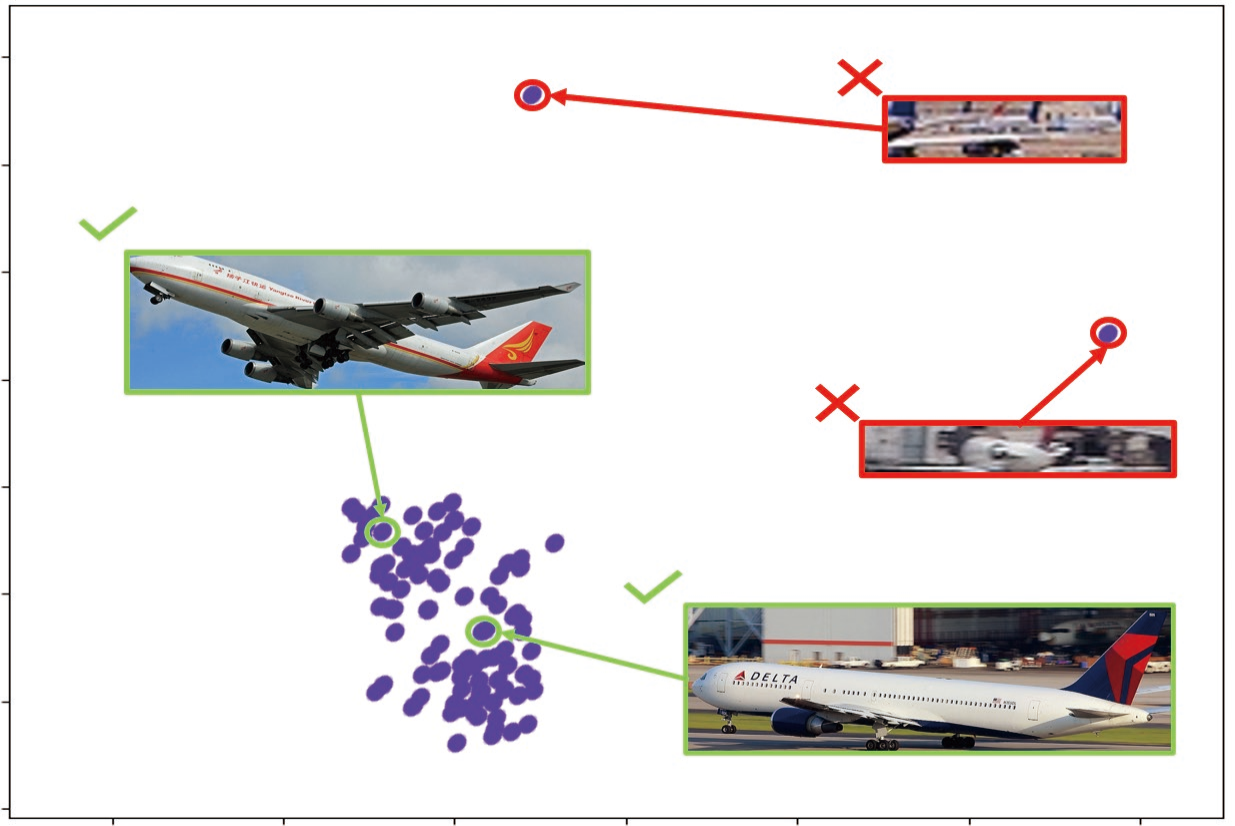}
    \caption{Quality of image prompts. Green indicates good image prompts, while red indicates poor ones.}
    \label{fg_good_bad_prompt}
\end{figure}

In prompt feature selection module, it is worth to highlight that the quality of image prompts have a significant impact on detection and segmentation. Directly integrating all the image prompts of the same category will lead to unstable performance due to the low information density of images. As shown in  Fig. \ref{fg_good_bad_prompt}, Despite most image prompt features exhibit good clustering properties, there still exist a few outliers caused by instances that are hard to recognize. These outliers will reduce the distinctiveness of the semantic information in image prompts. We observe that high-quality image prompt features within the same category tend to be highly similar, while low-quality ones always show significant differences. 

\begin{figure}[b]
    \centering
    \includegraphics[width=1.\linewidth]{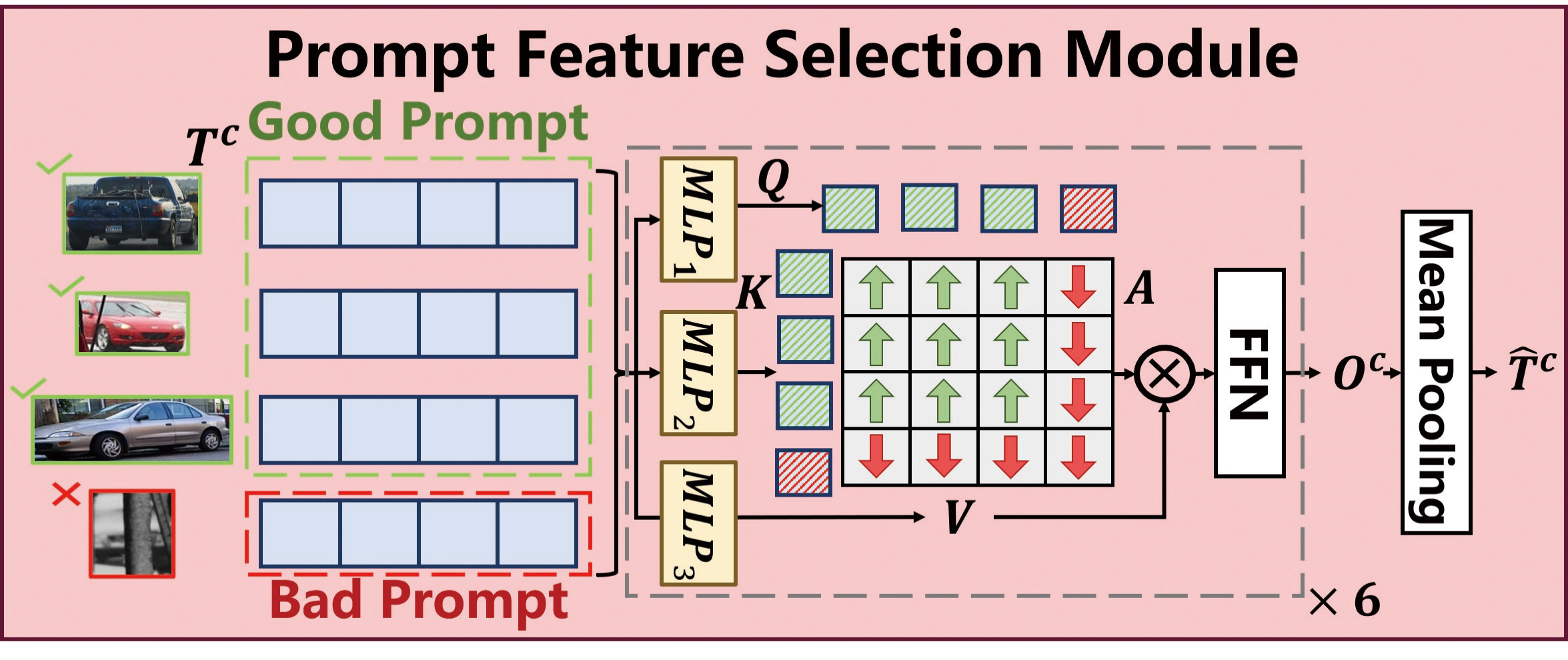}
    \caption{The overall framework of PFSM.}
    \label{fg_pfsm}
\end{figure}

Inspired by the above observation, we develop a prompt feature selection module based on self-attention (PFSM) to leverage the correlation between prompt features, reducing the impact of poor-quality image prompts, as shown in Fig \ref{fg_pfsm}. The overall process is illustrated in EQ. \ref{eq_pfem}, where $\theta$ represents the learnable parameters of $PFSM(\cdot)$:

\begin{equation}
    {\hat T^c} = PFSM({T^c},\theta ).
    \label{eq_pfem}
\end{equation}

In PFSM, we first use self-attention to calculate the correlation between the $N$ image prompts:

\begin{equation}
    Q = {MLP_1}({T^c}),K = {MLP_2}({T^c}),V = {MLP_3}({T^c}),
    \label{eq_qkv}
\end{equation}

\begin{equation}
    A = softmax({{Q{K^T}} \mathord{\left/
 {\vphantom {{Q{K^T}} {\sqrt {{D_1}} }}} \right.
 \kern-\nulldelimiterspace} {\sqrt {{D_1}} }}),
    \label{eq_attention}
\end{equation}

\noindent where $MLP(\cdot)$ is a fully connected network for feature dimension adjustment, and $Q,K,V \in \mathbb{R}^{N \times D_1}$ are the query, key, and value needed for self-attention. $A \in \mathbb{R}^{N \times N}$ is the correlation matrix between the $N$ image prompts. As analyzed earlier, the stronger the correlation with other prompt features, the more accurate the semantic information it contains. Conversely, weaker correlations suggest a higher likelihood of being an outlier. We assign higher weights to prompt features with more accurate semantic information:

\begin{equation}
    {O^c} = FFN(A{T^c}),
    \label{eq_matmul}
\end{equation}

\noindent where $O^c \in \mathbb{R}^{N \times D_2}$ represents the enhanced image prompt features, $FFN(\cdot)$ is a feed-forward layer, and $D_2$ is the transformed feature dimension. Finally, we apply average pooling to reduce the dimensionality of $O^c$ along the prompt quantity dimension:

\begin{equation}
    {\hat T^c} = MeanPooling({O^c} + Linear({T^c})),
    \label{eq_pooling}
\end{equation}

\noindent where $\hat{T}^c \in \mathbb{R}^{1 \times D_2}$ represents the final image prompt feature for category $c$. For all categories $\{1, 2, \dots, C\}$, $PFSM(\cdot)$ uses the same $\theta$ to obtain $\hat{T} = \{\hat{T}^1, \hat{T}^2, \dots, \hat{T}^C\}$.

\noindent \textbf{Vision Encoder.} To enhance the model's robustness to targets of different scales, we use a vision transformer backbone to construct the vision encoder, retaining the features from different layers as multi-scale features of the input image. The multi-scale features are defined as $F = \{f_1, \dots, f_L\}$, where $f_i$ represents the features from the $i$-th layer of the vison encoder.

\noindent \textbf{Transformer Encoder with Deep Fusion.} To reduce the difficulty of feature alignment, we fuse the enhanced image prompt features with the input image features by referencing the cross-modality interaction method from language-vision models. Specifically, we use a multi-scale deformable cross-attention~\cite{zhu2020deformable} to fuse the prompt features $\hat{T}$ with the multi-scale image features $F$, resulting in the object embeddings $\hat{F}$:

\begin{equation}
    \hat F = MSDeformAttn(F,\hat T).
    \label{eq_msda}
\end{equation}

\noindent \textbf{Region-Level Feature Alignment.} Inspired by the image-text alignment in the text prompt paradigm, we achieve region-level classification feature alignment between image prompts and predicted objects in the image prompt paradigm. Specifically, we directly compute the alignment scores $S \in {\mathbb{R}^{M \times C}}$ between the prompt features $\hat{T}$ and the object embeddings $\hat{F}$:

\begin{equation}
    S = \hat F \cdot \hat T,
    \label{eq_alignment}
\end{equation}

\noindent where $M$ is the predefined number of object embeddings. Finally, we use a transformer decoder to decode the object embeddings $\hat{F}$ into bounding box labels and mask labels.

\subsection{Training Strategy and Optimization Objective}

\begin{table*}[t]
  \centering
  \setlength{\extrarowheight}{-1mm}
  \setlength{\tabcolsep}{1mm}
  \begin{tabular}{@{}c|c|c|c|cccc|c@{}}
  \toprule
  \multirow{2}{*}{Method}        & \multirow{2}{*}{Training Data} & \multirow{2}{*}{\begin{tabular}[c]{@{}c@{}}Prompt\\ Paradigm\end{tabular}} & \begin{tabular}[c]{@{}c@{}}COCO\\ (out-domain)\end{tabular} & \multicolumn{4}{c|}{\begin{tabular}[c]{@{}c@{}}LVIS-1203\\ (out-domain)\end{tabular}} & \begin{tabular}[c]{@{}c@{}}ODinW-35\\ (out-domain)\end{tabular} \\ \cmidrule(l){4-9} 
                                 &                                &                                                                            & $AP^b$                                                          & $AP^b$                  & $AP_f^b$                 & $AP_c^b$                 & $AP_r^b$                 & $AP^b$                                                              \\ \midrule
  GLIP T~\cite{li2022grounded}                         & O365+GoldG+...                 & \multirow{4}{*}{Text}                                                      & 46.7                                                        & 17.2                & 25.5                & 12.5                & 10.1                & 19.6                                                            \\
  GLIP L~\cite{li2022grounded}                         & FourODs+GoldG+...              &                                                                            & 49.8                                                        & \ul{26.9}          & \textbf{35.4}       & \ul{23.3}          & \ul{17.1}          & \ul{23.4}                                                      \\
  Grounding DINO T~\cite{liu2023grounding}               & O365+GoldG+...                 &                                                                            & 48.4                                                        & -                   & -                   & -                   & -                   & 22.3                                                            \\
  Grounding DINO L~\cite{liu2023grounding}               & O365+GoldG+...                 &                                                                            & \ul{52.5}                                                  & -                   & -                   & -                   & -                   & \textbf{26.1}                                                   \\ \midrule
  DINOv T~\cite{li2024visual}                        & COCO+SA-1B                     & \multirow{2}{*}{Visual}                                                    & -                                                           & -                   & -                   & -                   & -                   & 14.9                                                            \\
  DINOv L~\cite{li2024visual}                        & COCO+SA-1B                     &                                                                            & -                                                           & -                   & -                   & -                   & -                   & 15.7                                                            \\ \midrule
  \textbf{MI Grounding-D (ours)} & O365                           & Image                                                                      & \textbf{53.7}                                               & \textbf{27.4}       & \ul{32.4}          & \textbf{25.8}       & \textbf{19.9}       & 21.5                                                            \\ \bottomrule
  \end{tabular}
  \caption{Object Detection with MI Grounding-D. Bold and underline denote the best and second-best results in each column. $AP^b$ represents the average precision for object detection. $AP_f^b$, $AP_c^b$, and $AP_r^b$ represent the average precision for frequent, common, and rare classes, respectively.}
  \label{tb_mg_d}
\end{table*}

\begin{table*}[t]
  \centering
  \setlength{\extrarowheight}{-1mm}
  \setlength{\tabcolsep}{1mm}
  \begin{tabular}{@{}c|c|c|cc|cc|cc@{}}
  \toprule
  \multirow{2}{*}{Method}        & \multirow{2}{*}{Training Data} & \multirow{2}{*}{\begin{tabular}[c]{@{}c@{}}Prompt\\ Paradigm\end{tabular}} & \multicolumn{2}{c|}{\begin{tabular}[c]{@{}c@{}}COCO\\ (in-domain)\end{tabular}} & \multicolumn{2}{c|}{\begin{tabular}[c]{@{}c@{}}ADE20K\\ (out-domain)\end{tabular}} & \multicolumn{2}{c}{\begin{tabular}[c]{@{}c@{}}SegInW\\ (out-domain)\end{tabular}} \\ \cmidrule(l){4-9} 
                                 &                                &                                                                            & $AP^m$                                   & $AP^b$                                   & $AP^m$                                     & $AP^b$                                    & $AP_{avg}^m$                                    & $AP_{med}^m$                                    \\ \midrule
  GLIPv2 H~\cite{zhang2022glipv2}                       & COCO+O365+...                  & \multirow{7}{*}{Text}                                                      & 48.9                                   & -                                      & -                                        & -                                       & -                                       & -                                       \\
  MaskCLIP L~\cite{ding2022open}                     & YFCC100M                       &                                                                            & -                                      & -                                      & 15.1                                     & 6.0                                     & 23.7                                    & -                                       \\
  FC-CLIP L~\cite{yu2024convolutions}                      & COCO                           &                                                                            & 44.6                                   & -                                      & \ul{16.8}                                     & -                                       & -                                       & -                                       \\
  OpenSeed T~\cite{zhang2023simple}                     & COCO+O365                      &                                                                            & 47.6                                   & 52.0                                   & 14.1                                     & 17.0                                    & 33.9                                    & 21.5                                    \\
  X-Decoder T~\cite{zou2023generalized}                    & COCO+CC3M+...                  &                                                                            & 40.5                                   & 43.6                                   & 9.8                                      & -                                       & 22.7                                    & 15.2                                    \\
  X-Decoder L~\cite{zou2023generalized}                    & COCO+CC3M+...                  &                                                                            & 46.7                                   & -                                      & 13.1                                     & -                                       & 36.1                                    & 38.7                                    \\
  OpenSeed L~\cite{zhang2023simple}                     & COCO+O365                      &                                                                            & \textbf{53.2}                          & \textbf{58.2}                          & 15.0                                     & \ul{17.7}                              & 36.1                                    & 38.7                                    \\ \midrule
  DINOv T~\cite{li2024visual}                        & COCO+SA-1B                     & \multirow{2}{*}{Visual}                                                    & 41.5                                   & 45.2                                   & 11.4                                     & 12.8                                    & 39.5                                    & \ul{41.6}                              \\
  DINOv L~\cite{li2024visual}                        & COCO+SA-1B                     &                                                                            & \ul{50.4}                             & 54.2                                   & 15.1                               & 14.3                                    & \ul{40.6}                              & \textbf{44.6}                           \\ \midrule
  \textbf{MI Grounding-S (ours)} & COCO+LVIS                      & Image                                                                      & 46.1                                   & \ul{54.7}                             & \textbf{21.0}                            & \textbf{25.3}                           & \textbf{46.9}                           & 41.3                                    \\ \bottomrule
  \end{tabular}
  \caption{Instance Segmentation with MI Grounding-S. Bold and underline denote the best and second-best results in each column.  $AP^b$ represents the average precision for object detection.  $AP^m$ represents the average precision for instance segmentation.}
  \label{tb_mg_s}
\end{table*}

\textbf{Image Prompt Training Strategy.} To enhance the model's generalization ability, we use a random image prompt strategy. During training, we randomly sample $N$ cropped instance images as image prompts for each category, updating them every iteration. The random sampling allows the model to adapt to cross-domain image prompts. The frequent updates help the model learn and adjust to more complex prompts. It's important to note that during testing, the model also uses only instance images cropped from the training set as prompts. This not only prevents data leakage from the test set but also demonstrates the model's generalization ability to cross-domain image prompts.

\noindent \textbf{Optimization Objective.} Since our model directly predicts the target's class, box, and mask in an end-to-end manner, the loss function $\mathcal{L}$ of MI Grounding consists of classification loss $\mathcal{L}_{class}$, localization losses $\mathcal{L}_{L1}$ and $\mathcal{L}_{GIoU}$, and segmentation loss $\mathcal{L}_{mask}$:

\begin{equation}
    \mathcal{L} = {\mathcal{L}_{class}} + {\mathcal{L}_{L1}} + {\mathcal{L}_{GIoU}} + {\mathcal{L}_{mask}}.
    \label{eq_loss}
\end{equation}

For the classification loss, we use a contrastive loss~\cite{radford2021learning} to calculate the difference between the predicted target and the image prompt features for open-set classification. For the localization loss, we apply L1 loss~\cite{ren2015faster} for regressing the bounding box coordinates and GIoU loss~\cite{rezatofighi2019generalized} to enhance convergence stability. In the segmentation loss, $\mathcal{L}_{mask}$ is a cross-entropy loss for mask segmentation.

\section{Experiments}

\subsection{Datasets and Settings}
In our experiments, we provide two sets of model parameters: MI Grounding-S for segmentation and MI Grounding-D for object detection. In MI Grounding-S, we use only the COCO~\cite{lin2014microsoft} and LVIS~\cite{gupta2019lvis} datasets for joint training and test on the COCO, ADE20K~\cite{zhou2017scene}, and SegInW~\cite{zou2023generalized} datasets. In MI Grounding-D, we use only the Objects365~\cite{shao2019objects365} dataset for training and test on the COCO, LVIS, and ODinW~\cite{li2022grounded} datasets. In both MI Grounding-S and MI Grounding-D, we use ViT-L as the vision backbone. We use 8 as the number of image prompts in our method, as discussed in the ablation study.

\subsection{Comparison to Prior Works}
To explore the generalization ability of the image prompt paradigm and MI Grounding, we test our model on multiple datasets across different domains. It's important to note that we train MI Grounding-D on Objects365 for 32 A100 days and MI Grounding-S on COCO+LVIS for 16 A100 days. Our training data and duration are significantly less than most methods in Table. \ref{tb_mg_d} and Table. \ref{tb_mg_s}. For example, GLIP L is trained on the FourODs GoldG, and Cap24M datasets for 600 V100 days~\cite{li2022grounded}.

\noindent \textbf{Object Detection with MI Grounding-D.} In Table. \ref{tb_mg_d}, we test on well-established benchmarks, including common object detection datasets like COCO, long-tailed datasets like LVIS, and complex cross-domain datasets like ODinW. MI Grounding-D demonstrates strong performance in out-of-domain scenarios. Notably, MI Grounding-D leads by 2.8\% in AP for rare categories on LVIS, further highlighting the generalization ability of the image prompt paradigm.

\noindent \textbf{Instance Segmentation with MI Grounding-S.} As shown in Table. \ref{tb_mg_s}, we test MI Grounding-S on multiple datasets under both in-domain and out-domain conditions. Notably, in out-domain scenarios, MI Grounding-S achieves a significant advantage of 4.2\% on ADE20K and 6.3\% on SegInW. SegInW is a complex cross-domain dataset containing 25 different sub-datasets, and the performance advantage on this dataset underscores the generalization ability of the image prompt paradigm.

\subsection{Ablation}

\textbf{Effectiveness of Prompt Feature Selection Module.} To demonstrate the effectiveness of the prompt feature selection module (PFSM) based on self-attention in the image prompts selection encoder, we replace it with three other modules: a fully connected network, a convolutional neural network, and mean pooling. The results in Table. \ref{tb_pfsm} are obtained from training and testing only on COCO. In the fully connected network and convolutional network, we use supervised neural networks to reduce the dimensionality of the $N$ image prompt features. In the mean pooling, we directly take the mean of the $N$ image prompt features to obtain the final prompt feature. As shown in Table. \ref{tb_pfsm}, our proposed PFSM proves to be the most effective among the various strategies.

\begin{table}[h]
    \setlength{\extrarowheight}{-1mm}
    \centering
    \begin{tabular}{@{}c|cccc@{}}
    \toprule
    \multirow{2}{*}{\begin{tabular}[c]{@{}c@{}}Prompt Feature\\ Calculation\end{tabular}} & \multicolumn{4}{c}{COCO}                                                           \\ \cmidrule(l){2-5} 
                                                                                          & $AP^b$           & \multicolumn{1}{c|}{$AP^b_{med}$}       & $AP^m$           & $AP^m_{med}$       \\ \midrule
    Mean Pooling                                                                          & 56.5           & \multicolumn{1}{c|}{60.9}           & 48.4           & 52.6           \\
    FC                                                                                    & 54.5          & \multicolumn{1}{c|}{59.9}          & 47.2          & 51.2          \\
    CNN                                                                                   & \ul{58.0}          & \multicolumn{1}{c|}{\ul{62.4}}          & \ul{49.7}          & \ul{53.4}          \\
    \textbf{PFSM}                                                                         & \textbf{61.7} & \multicolumn{1}{c|}{\textbf{67.2}} & \textbf{53.1} & \textbf{57.8} \\ \bottomrule
    \end{tabular}
    \caption{Ablation study of PFSM. FC represents the fully connected network, and CNN represents the convolutional neural network.}
    \label{tb_pfsm}
\end{table}

\noindent \textbf{Impact of Image Prompt Update Frequency.} Even within the same category, instance images can be highly diverse. To help the model adapt to this diversity, we increase the frequency of image prompt updates, allowing the model to learn from a wider range of image prompts during training. As shown in Table. \ref{tb_update}, we demonstrate the impact of update frequency on model performance. We gradually increase the update frequency from once every 200 iterations to once per iteration, and the model's performance improve accordingly. Finally, we set the model to update the image prompts once per iteration.

\begin{table}[h]
    \setlength{\extrarowheight}{-1mm}
    \centering
    \begin{tabular}{@{}c|cccc@{}}
    \toprule
    \multirow{2}{*}{\begin{tabular}[c]{@{}c@{}}Prompt Update\\ Frequency\end{tabular}} & \multicolumn{4}{c}{COCO}                                                           \\ \cmidrule(l){2-5} 
                                                                                       & $AP^b$           & \multicolumn{1}{c|}{$AP^b_{med}$}       & $AP^m$           & $AP_m^{med}$       \\ \midrule
    200 Iterations                                                                     & 57.9          & \multicolumn{1}{c|}{63.4}          & 49.7          & 54.4          \\
    150 Iterations                                                                     & 58.8          & \multicolumn{1}{c|}{64.1}          & 50.6          & 55.3          \\
    100 Iterations                                                                     & \ul{60.58}   & \multicolumn{1}{c|}{\ul{65.87}}   & \ul{52.2}    & \ul{57.0}    \\
    50 Iterations                                                                      & 59.2          & \multicolumn{1}{c|}{64.5}          & 50.7          & 55.2          \\
    \textbf{1 Iteration}                                                               & \textbf{61.7} & \multicolumn{1}{c|}{\textbf{67.2}} & \textbf{53.1} & \textbf{57.8} \\ \bottomrule
    \end{tabular}
    \caption{Ablation study of image prompt update frequency.}
    \label{tb_update}
\end{table}

\noindent \textbf{Impact of Image Prompt Quantity.} As previously mentioned, due to the low information density of images, a single image prompt often fails to fully convey the semantic information of the target class. As the number of image prompts increases, the semantic representation of the class becomes more complete, leading to higher-quality open-set visual perception. In Table. \ref{tb_ip_quantity}, we perform an ablation study on the number of image prompts. As the number of image prompts increases, the model's performance gradually improves. However, when the number of image prompts exceeds 8, there is no significant performance gain. More seriously, with the number of image prompt increases, the computational cost of the model increases. Therefore, the optimal setting of image prompt quantity is 8.

\begin{table}[h]
  \setlength{\extrarowheight}{-1mm}
  \centering
  \begin{tabular}{@{}c|cccc@{}}
  \toprule
  \multirow{2}{*}{\begin{tabular}[c]{@{}c@{}}Image Prompt\\ Quantity\end{tabular}} & \multicolumn{4}{c}{COCO}                                                           \\ \cmidrule(l){2-5} 
                                                                                   & $AP^b$           & \multicolumn{1}{c|}{$AP^b_{med}$}       & $AP^m$           & $AP^m_{med}$       \\ \midrule
  1                                                                                & 44.6          & \multicolumn{1}{c|}{48.5}          & 38.5          & 41.4          \\
  3                                                                                & 56.2          & \multicolumn{1}{c|}{61.4}          & 48.3          & 52.9          \\
  5                                                                                & 59.4          & \multicolumn{1}{c|}{64.5}          & 51.3          & 55.6          \\
  \textbf{8}                                                                       & \textbf{61.7} & \multicolumn{1}{c|}{\textbf{67.2}} & \textbf{53.1} & \textbf{57.8} \\
  12                                                                               & 61.4          & \multicolumn{1}{c|}{66.6}          & 52.7          & 57.2          \\
  20                                                                               & \ul{61.6}    & \multicolumn{1}{c|}{\ul{66.9}}    & \ul{52.8}    & \ul{57.4}    \\ \bottomrule
  \end{tabular}
  \caption{Ablation study of image prompt quantity. Model performance improves as the number of image prompts increases. After reaching 8 prompts, there is no significant further improvement.}
  \label{tb_ip_quantity}
\end{table}

\subsection{ADR50K Dataset}

\begin{table*}[t]
  \setlength{\extrarowheight}{-1mm}
  \centering
  \footnotesize
  \begin{tabular}{@{}c|c|cccc|cccc@{}}
  \toprule
  \multirow{2}{*}{Method}      & \multirow{2}{*}{\begin{tabular}[c]{@{}c@{}}Prompt\\ Paradigm\end{tabular}} & \multicolumn{4}{c|}{Object Detection}                                 & \multicolumn{4}{c}{Instance Segmentation}                             \\ \cmidrule(l){3-10} 
                               &                                                                            & $AP_{50}$          & $AP_{sink}$       & $AP_{shrinkage}$ & $AP_{porosity}$   & $AP_{50}$          & $AP_{sink}$       & $AP_{shrinkage}$ & $AP_{porosity}$   \\ \midrule
  Grounding DINO L             & Text                                                                       & 43.4          & 43.1          & 18.2                  & 17.5          & -             & -             & -                     & -             \\
  DINOv L                      & Visual                                                                     & 16.5          & 21.6          & 1.403                 & 0.1           & 15.4          & 13.8          & 1.0                   & 0.11          \\ \midrule
  \textbf{MI Grounding (ours)} & Image                                                                      & \textbf{50.2} & \textbf{49.6} & \textbf{19.4}         & \textbf{25.8} & \textbf{50.4} & \textbf{45.0} & \textbf{15.8}         & \textbf{27.9} \\ \bottomrule
  \end{tabular}
  \caption{Comparative experiments on ADR50K. $AP_{50}$ represents the average precision for all categories at an IoU threshold of 0.50. $AP_{sink}$, $AP_{shrinkage}$, and $AP_{porosity}$ represent the average precision for ``sink", ``shrinkage porosity", and ``porosity", respectively.}
  \label{tb_adr50k_ex}
\end{table*}

To further demonstrate the advantages of the image prompt paradigm, we create the Automatic Defect Recognition dataset (ADR50K). In ADR50K, we collect more than 50,000 X-ray images of defect inspections. We provide classification annotations for three types of defects using specialized terminology: ``sink", ``shrinkage porosity", and ``porosity". Additionally, we provide detection and segmentation annotations for all defects. The relevant details of the ADR50K dataset are shown in Table. \ref{tb_adr50k}.

Unlike conventional object detection and segmentation datasets, the detection targets and class names in the ADR50K dataset are highly specialized and even misleading. As shown in Figure 5, we present some example images of the three defect types in the dataset. The ADR50K dataset poses three main challenges for open-set visual perception methods. \textbf{(1) Difficulty in aligning category names with instances.} In typical scenarios, ``sink" usually refers to a basin, commonly found in kitchens or bathrooms, where water is supplied through a faucet and drains away. However, in the ADR50K dataset, ``sink" refers to a slender indentation that is completely different from a basin. \textbf{(2) Confusion between category names.} The terms ``shrinkage porosity" and ``porosity" seem to be same category as their textual name look similar with each other, but they refer to entirely different types of defects. In the ADR50K dataset, ``shrinkage porosity" refers to a sheet-like shallow depression, while "porosity" refers to small round pits. \textbf{(3) Confusion with background images.} The images in the ADR50K dataset contain numerous shadows caused by overlapping structural components, which are not defects. These shadows can easily be mistaken for defects that need to be detected.

\begin{table}[t]
  \setlength{\extrarowheight}{-1mm}
  \centering
  \footnotesize
  \begin{tabular}{@{}c|c|ccc@{}}
  \toprule
  \multirow{2}{*}{\begin{tabular}[c]{@{}c@{}}Data\\ Split\end{tabular}} & \multirow{2}{*}{\begin{tabular}[c]{@{}c@{}}Number of\\ Images\end{tabular}} & \multicolumn{3}{c}{\begin{tabular}[c]{@{}c@{}}Number of\\ Instances\end{tabular}} \\ \cmidrule(l){3-5} 
                                                                        &                                                                             & \multicolumn{2}{c|}{Categories}                        & Total                    \\ \midrule
  \multirow{3}{*}{Train}                                                & \multirow{3}{*}{51572}                                                      & ``sink"                & \multicolumn{1}{c|}{167886}   & \multirow{3}{*}{196673}  \\
                                                                        &                                                                             & ``shrinkage porosity"   & \multicolumn{1}{c|}{24320}    &                          \\
                                                                        &                                                                             & ``porosity"             & \multicolumn{1}{c|}{4467}     &                          \\ \midrule
  \multirow{3}{*}{Test}                                                 & \multirow{3}{*}{7696}                                                       & ``sink"                & \multicolumn{1}{c|}{20946}    & \multirow{3}{*}{24437}   \\
                                                                        &                                                                             & ``shrinkage porosity"   & \multicolumn{1}{c|}{2955}     &                          \\
                                                                        &                                                                             & ``porosity"             & \multicolumn{1}{c|}{536}      &                          \\ \bottomrule
  \end{tabular}
  \caption{ADR50K Dataset Details. We allocated approximately 10\% of the data to the test set, with the remaining data used for the training set.}
  \label{tb_adr50k}
\end{table}

\begin{figure}[h]
    \centering
    \includegraphics[width=1.\linewidth]{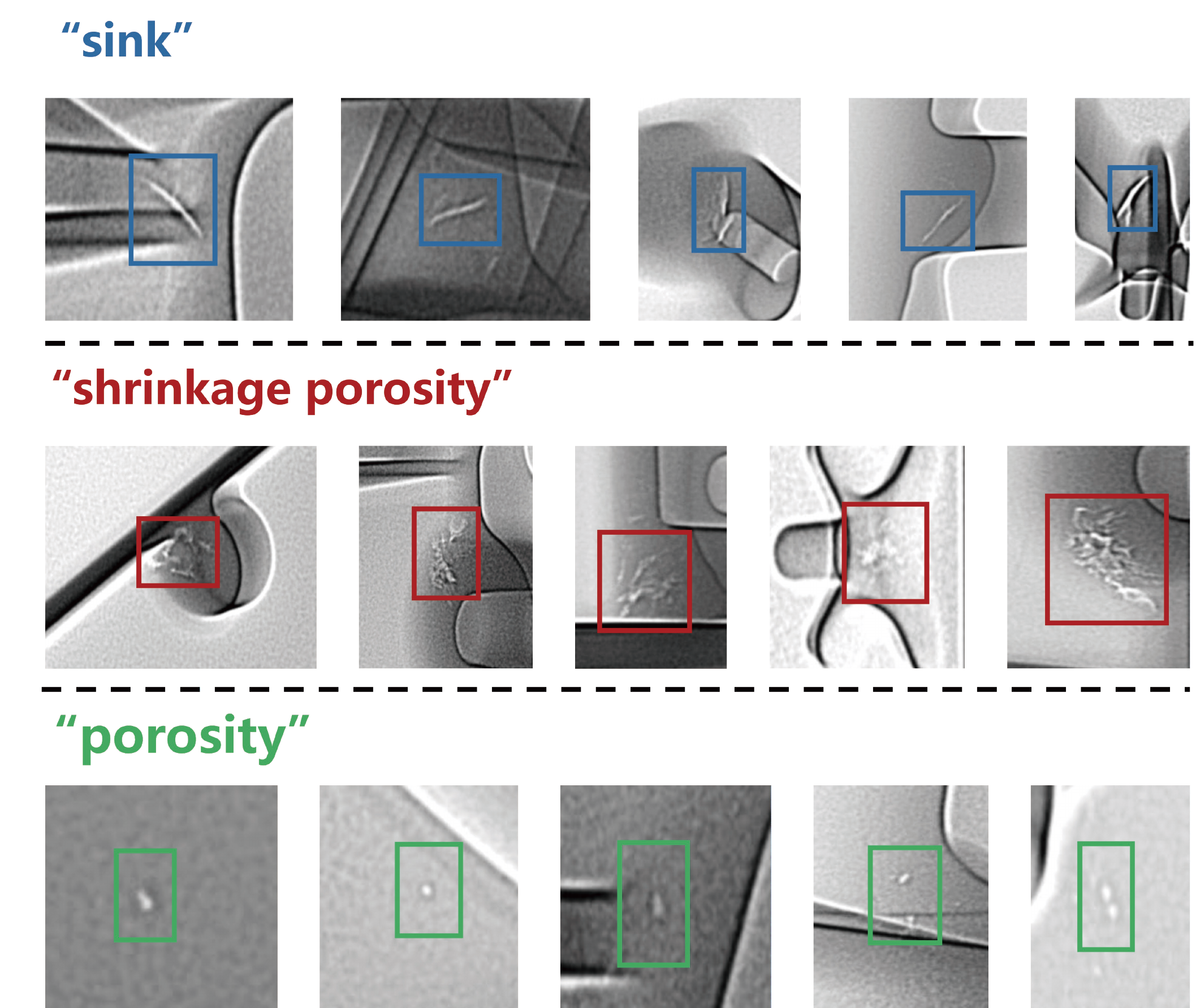}
    \caption{Examples of specialized categories in ADR50K dataset. The text denotes the category name of defects, and the areas within bounding boxs denote the corresponding category region.}
    \label{fg_adr50k}
\end{figure}

We compared our proposed image prompt method with text prompt and visual prompt methods on the ADR50K dataset. As shown in Table. \ref{tb_adr50k_ex}, MI Grounding outperforms Grounding DINO L~\cite{liu2023grounding} and DINOv L~\cite{li2024visual} in both object detection and instance segmentation tasks. In MI Grounding, we use instance images of defects as prompts instead of potentially misleading text. Additionally, unlike visual prompt methods, MI Grounding in the image prompt paradigm does not require a separate prompt for each target instance.

\section{Conclusion}

In this paper, we introduce a novel visual perception paradigm called the Image Prompt Paradigm. Unlike existing text and visual prompts, this paradigm uses a few image instances as prompts, enabling it to understand specialized categories which are challenging to describe with text in a single-stage and non-interactive manner. To support this new paradigm, we present a framework named MI Grounding. MI Grounding utilizes multiple image prompts to perform Open-Set Object Detection and Open-Set Segmentation, and it includes an image prompt selection encoder designed to choose and integrate high-quality prompts effectively. Our approach achieves competitive performance across several datasets when compared to mainstream methods in Open-Set Object Detection and Open-Set Segmentation, demonstrating the effectiveness of the proposed iamge prompt paradigm. To further validate the superiority, we developed a specialized ADR50K dataset, which comprises an extensive collection of X-ray defect detection data. Experimental results show that our approach significantly enhances performance on this specialized dataset.

\bibliography{aaai25}

\end{document}